\documentclass{article}

\usepackage{microtype}
\usepackage{graphicx}
\usepackage{subcaption}
\usepackage{booktabs} 

\usepackage{hyperref}

\usepackage[preprint]{icml2026}

\usepackage{amsmath}
\usepackage{amssymb}
\usepackage{mathtools}
\usepackage{amsthm}

\usepackage[capitalize,noabbrev]{cleveref}

\theoremstyle{plain}

\theoremstyle{definition}

\theoremstyle{remark}

\usepackage[textsize=tiny]{todonotes}

\usepackage{makecell}
\usepackage[table]{xcolor}
\usepackage{amssymb}
\usepackage[table]{xcolor}
\definecolor{lred}{RGB}{255,220,220}

\makeatletter
\newcommand{\appendixtableofcontents}{
  \section*{Appendix Table of Contents}
  \@starttoc{apx}
}

\makeatother
\usepackage{lineno}
\usepackage{subcaption}
\definecolor{darkblue}{rgb}{0, 0, 0.5}
\hypersetup{colorlinks=true, citecolor=darkblue, linkcolor=darkblue, urlcolor=darkblue}
\usepackage{multirow}
\usepackage{lineno}
\usepackage{mdframed}
\usepackage{fancyvrb}
\definecolor{darkblue}{rgb}{0, 0, 0.5}
\hypersetup{colorlinks=true, citecolor=darkblue, linkcolor=darkblue, urlcolor=darkblue}
 
\usepackage{listings}
\usepackage{xcolor}
\definecolor{darkblue}{rgb}{0, 0, 0.5}
\hypersetup{colorlinks=true, citecolor=darkblue, linkcolor=darkblue, urlcolor=darkblue}

\icmltitlerunning{Training Language Models to Cooperate with Inference-Time Controllers}

\begin{document}

\twocolumn[
  \icmltitle{Training Language Models to Cooperate with Inference-Time Controllers}

  \icmlsetsymbol{equal}{*}

  \begin{icmlauthorlist}
    \icmlauthor{Moumita Choudhury}{umass}
    \icmlauthor{Vanshaj Khattar}{merl}
    \icmlauthor{Jing Liu}{merl}
    \icmlauthor{Toshiaki Koike-Akino}{merl}
    \icmlauthor{Ankush Chakrabarty}{merl}
    \icmlauthor{Shlomo Zilberstein}{umass}
    \icmlauthor{Ye Wang}{merl}
   
  \end{icmlauthorlist}

  \icmlaffiliation{umass}{
College of Information and Computer Sciences,
University of Massachusetts Amherst,
Amherst, Massachusetts, USA
}

\icmlaffiliation{merl}{
Mitsubishi Electric Research Laboratories,
Cambridge, Massachusetts, USA
}

  \icmlcorrespondingauthor{Moumita Choudhury}{amchoudhury@umass.edu}

  \icmlkeywords{Machine Learning, ICML}

  \vskip 0.3in
]

\printAffiliationsAndNotice{} 

\begin{abstract}
Large language model (LLM) performance increasingly depends not only on the base model, but also on the inference-time controller used to organize reasoning. Existing post-training methods, however, typically optimize for a single fixed interaction pattern, despite real deployments relying on diverse controllers such as Chain-of-Thought, self-consistency, debate, planning, and verification pipelines. This creates a training--deployment mismatch and limits transfer to new workflows. We introduce \textbf{CALM} (\textbf{C}ontroller-\textbf{A}ware \textbf{L}anguage \textbf{M}odels), a post-training framework that explicitly places controllers in the training loop. We formulate controller-aware post-training as multi-task reinforcement learning over controller-induced interaction protocols, where controllers are compositions of reusable local reasoning modules. This structure also induces a module-level decomposition of mixed-controller training under a turn-level GRPO objective, enabling a systematic study of controller and module-aware training strategies. We evaluate CALM on held-out controller compositions and broader controller shifts, showing that controller-aware post-training improves generalization across inference-time workflows beyond single-controller optimization.
\end{abstract}

\section{Introduction}
Artificial intelligence systems based on large language models (LLMs) are rapidly evolving from standalone predictors into general-purpose agents capable of reasoning, planning, and interacting with complex environments \citep{brown2020language, chowdhery2022palm, touvron2023llama}. Rather than relying on a single query to the model, modern LLM systems increasingly operate through \emph{structured inference-time procedures} that allocate additional computation across multiple calls, intermediate reasoning steps, sampling, aggregation, search, or interaction with tools and memory. This shift has made the performance of LLM agents depend not only on the capability of the underlying model, but also on how inference is organized and orchestrated at test time.

\begin{figure*}[!htb]
    \centering
    \includegraphics[width=0.95\linewidth]{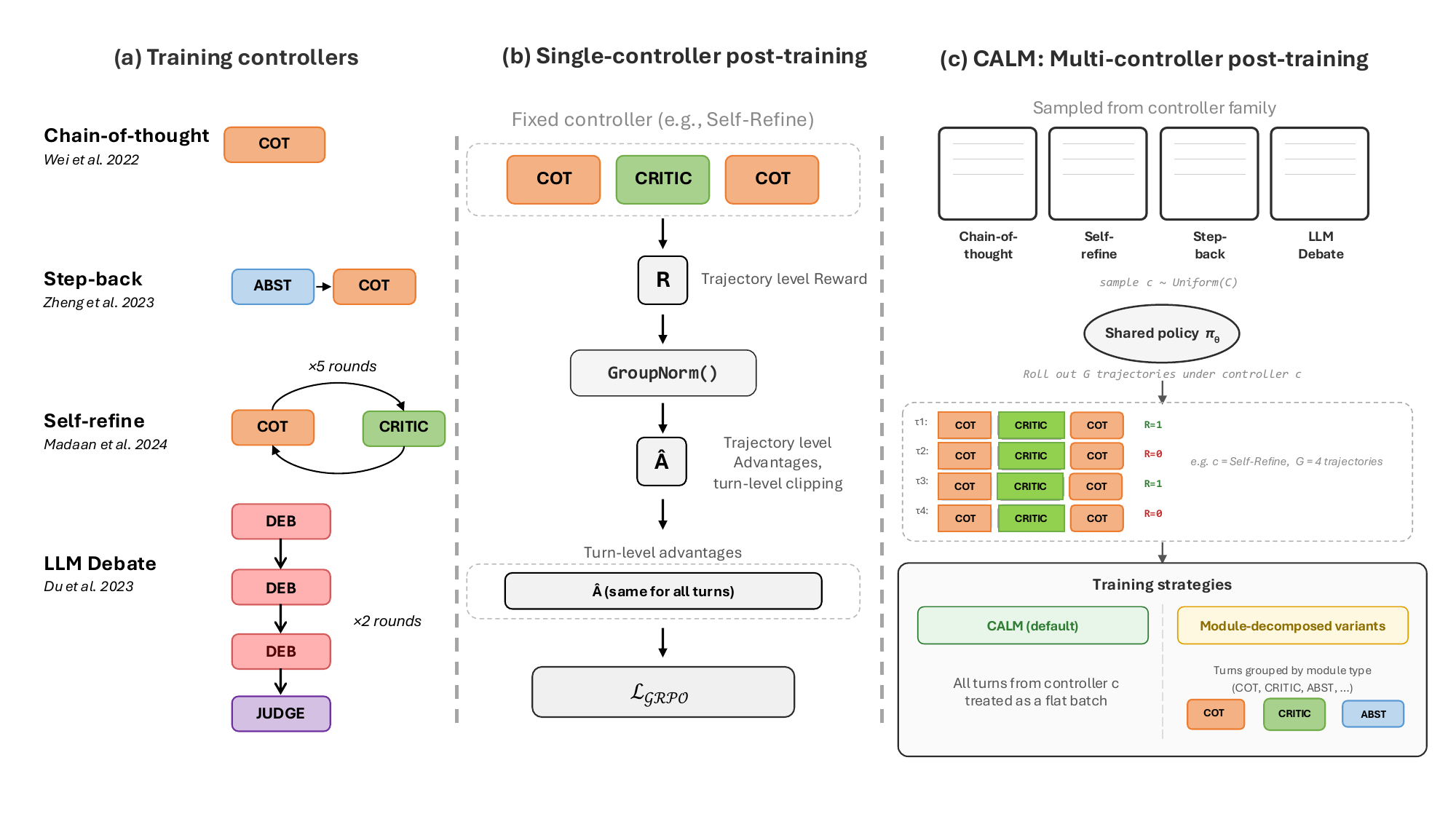}
    \caption{Overview of CALM. (a) Training controllers are compositions of reusable modules (COT, CRITIC, ABSTRACTOR, DEBATER, JUDGE, etc.) drawn from established inference-time methods~~\citep{hu2025adas}. Each controller consists of diverse set of roles and could follow a number of turns with or without memory.  (b) Single-controller post-training baseline, where a fixed controller generates trajectories. We use a multi-turn variation of the GRPO~\cite{wan2025rema} objective. (c) CALM's multi-controller post-training approach, which samples a controller uniformly at each iteration, rolls out multiple trajectories under a shared policy, and trains using either a flat batch of all turns or module-decomposed variants grouped by turn type.}
\label{fig:fig1}
\end{figure*}
A growing family of inference-time procedures has emerged to enhance LLM reasoning and agency. Foundational methods include Chain-of-Thought prompting \citep{wei2022chainofthought}, Self-Consistency \citep{wang2023selfconsistency}, Tree-of-thought, and Algorithm-of-thought \citep{yao2023tree, sel2024algorithm}, while more structured pipelines integrate Self-refinement \citep{madaan2024selfrefine, shinn2023reflexion}, Multi-agent debate \citep{du2023debate}, Tool-augmented reasoning \citep{schick2023toolformer}, and workflows that combine planning, memory, and external actions \citep{yao2022react}. We refer to such inference-time procedures as \emph{controllers}: a controller specifies how the model is queried over one or more calls, including the number of turns, the active role at each turn, and any aggregation, verification, or search logic. As controllers are now becoming sophisticated enough to be automatically searched and optimized \citep{hu2025adas, saad2024archon}, the performance still depends on the base model which is typically post-trained for a single fixed interaction pattern. For example, prior work discusses fine-tuning strategies for specific inference-time techniques such as Best-of-N sampling \citep{chow2025bon}. This mismatch is consequential: modern deployments rely on a \emph{family} of controllers that often share recurring local reasoning behaviors, such as CoT generation, verification, planning, or critique, while differing in call structure, role composition, and number of turns. A model trained on only one workflow can over-specialize to that controller and fail to transfer when familiar reasoning behaviors appear in novel compositions. This raises a natural question: \emph{how should the base model itself be post-trained to cooperate effectively with a \textbf{family} of controllers, rather than a single fixed one?}

To address this question, we introduce \textbf{CALM} (\textbf{C}ontroller-\textbf{A}ware \textbf{L}anguage \textbf{M}odels), a post-training framework that explicitly places controllers in the training loop. We formulate controller-aware post-training as a \emph{multi-task RL problem over controllers}. Each controller defines a Markov Decision Process (MDP), and training proceeds over a distribution of such MDPs rather than a single fixed one. Furthermore, controllers are not independent tasks; they share recurring local components such as CoT generation, verification, planning, critique, etc., while differing in how these components are arranged and ordered. We call these shared components \emph{modules}. A controller is therefore a \emph{composition} of reusable modules, and the central hypothesis of CALM is that if the model learns these modules well, it can generalize to unseen controller compositions. Our controllers are implemented as executable Python programs, following the code-as-controller paradigm of ADAS~\citep{hu2025adas} where each controller is a function that sequences module calls, routes outputs between turns, and handles branching or iteration logic, making the full controller inventory transparent, reproducible, and automatically extensible to new compositions.

Since controllers share reusable modules, but differ in how those modules are arranged, CALM opens a space of module-level training strategies that does not exist in single-controller post-training.
We study four variants that decompose the multi-controller loss by module type under a turn-level GRPO objective~\citep{shao2024deepseekmath, wan2025rema}: flat mixing of all turns, uniform weighting across module types, adaptive gradient-norm balancing, and single-role-per-step training.
Because controllers differ in turn count and role composition, some modules naturally produce larger gradient updates than others; the variants explore whether explicitly accounting for this imbalance improves cross-controller generalization. We aim to map the design space that the compositional controller view opens and to identify which axes matter most for cross-controller transfer.

Our experiments are designed to measure whether CALM truly learns reusable controller-aware reasoning rather than simply memorizing training workflows. We evaluate both on held-out controller compositions built from familiar reasoning primitives and on more challenging controller shifts that alter the inventory of local roles or interfaces. 
The former isolates whether CALM can recombine known reasoning behaviors in novel workflows, while the latter measures whether the shared policy remains robust under broader structural shift. 
We further evaluate across mathematical reasoning tasks to assess whether the findings generalize beyond a single task distribution.

In summary, this paper makes the following contributions:

1) formulating controller-aware LLM post-training as a \emph{multi-task RL problem over controller-induced MDPs}, which aims to learn a single policy that generalizes across controllers rather than optimizing for one fixed workflow,
2) proposing CALM as a framework for multi-controller training, and systematically investigate a range of module-level training strategies, including mixed module training, uniform weighting, adaptive gradient-norm balancing, and single-role-per-step training to understand the impact of different strategies, and
3) evaluating CALM on held-out controller compositions and broader controller shifts across mathematical reasoning tasks, separating transfer to new workflows built from familiar reasoning primitives from robustness to more substantial controller variation.
Our experiments show that multi-controller training consistently outperforms single-controller baselines, which overfit to their native controller and degrade on unseen compositions, with improvements holding under both compositional recombination and broader controller shift across math reasoning tasks.

\section{Preliminaries}
 
We briefly review the RL objectives that CALM builds on.
RL for language models frames decoding as a token-level MDP~\citep{wang2024openr} with objective $\mathcal{J}(\theta) = \mathbb{E}[R(\mathbf{y}, \mathbf{y}^*)]$.
GRPO~\citep{shao2024deepseekmath} optimizes this via a clipped importance-weighted objective over groups of $G$ sampled responses:
\begin{equation}
\small
\begin{aligned}
\mathcal{J}(\theta)
= & \mathbb{E}\Bigg[
\frac{1}{G}
\sum_{i=1}^{G}
\frac{1}{|\mathbf{y}_i|}
\sum_{j=1}^{|\mathbf{y}_i|}
\Bigg(
\min\Big(
r_{i,j}(\theta)\hat{A}_{i,j}, \\
&\mathrm{clip}\big(r_{i,j}(\theta), 1{-}\epsilon, 1{+}\epsilon\big)\hat{A}_{i,j}
\Big) - \beta D_{\mathrm{KL}}\big(\pi_\theta \,\Vert\, \pi_{\mathrm{ref}}\big)
\Bigg)
\Bigg]
\end{aligned}
\label{eq:grpo_obj}
\end{equation}
where $r_{i,j}(\theta) = \pi_\theta(y_{i,j} \mid \mathbf{x}, \mathbf{y}_{i,<j})\,/\,\pi_{\theta_{\mathrm{old}}}(y_{i,j} \mid \mathbf{x}, \mathbf{y}_{i,<j})$ is the token-level importance ratio and $\hat{A}_{i,j} = (R_i - \mu) / \sigma$ is the group-normalized advantage.
 
\paragraph{Turn-level GRPO for multi-turn interaction.}
Standard GRPO applies clipping at the token level, which introduces length bias in multi-turn trajectories and can encourage excessively long single-turn responses.
ReMA~\citep{wan2025rema} addresses this with a \emph{turn-level ratio}: the importance ratio for turn $t$ is averaged over all tokens within that turn,
$r_{i,t}(\theta) = \frac{1}{|\mathbf{y}_{i,t}|}\sum_j r_{i,t,j}(\theta)$,
and clipping is applied at the turn level rather than the token level.
The resulting trajectory-level objective is:
\begin{equation}
\small
\begin{aligned}
\mathcal{J}(\theta)
&= \mathbb{E}\Bigg[
\frac{1}{G}\sum_{i=1}^{G}
\frac{1}{T_i}\sum_{t=1}^{T_i}
\frac{1}{|\mathbf{y}_{i,t}|}\sum_{j=1}^{|\mathbf{y}_{i,t}|}
\Bigg(
\min\Big(
r_{i,t}(\theta)\hat{A}_{i,t,j}, \\
&\qquad \mathrm{clip}\big(r_{i,t}(\theta), 1{-}\epsilon, 1{+}\epsilon\big)\hat{A}_{i,t,j}
\Big)
- \beta D_{\mathrm{KL}}\big(\pi_\theta \,\Vert\, \pi_{\mathrm{ref}}\big)
\Bigg)
\Bigg]
\end{aligned}
\label{eq:turn_level_grpo}
\end{equation}
CALM adopts this turn-level formulation and extends it to the multi-controller setting, where controllers differ substantially in turn count and role composition.

\section{Problem Formulation}
\label{sec:formulation}
We formalize CALM as a multi-task reinforcement learning problem over a family of
controller-induced interaction protocols.

\paragraph{Controllers and modules.}
A \emph{controller} $c$ is a structured inference-time procedure that orchestrates one or more LLM calls to solve a given problem.
Each controller defines a set of \emph{modules} $\mathcal{M}_c = \{m_1, \dots, m_K\}$ and a \emph{control flow} $f_c$.
A module $m_k = (\rho_k, \kappa_k)$ pairs a \textbf{role} $\rho_k$ (e.g., \texttt{Chain-of-Thought Agent}, \texttt{Critic Agent}) with a \textbf{turn interface} $\kappa_k$ that specifies a role-specific instruction prompt and a structured output schema (e.g., \texttt{[thinking, answer]}, \texttt{[feedback, correct]}, etc.).
The control flow $f_c$ is a program that determines the invocation order of modules, routes the output of one module as input to the next, and may include, but not limited to, non-LLM logic such as conditional branching or iteration, tool calls, mathematical operators, etc.
For example, Self-Refine~\citep{madaan2024selfrefine} defines three modules (a \texttt{CoT Agent}, a \texttt{Critic Agent}, and the same \texttt{CoT Agent} re-invoked with a reflection prompt) and a control flow that loops the critique--revise cycle up to $N$ times or until the critic signals correctness.

\paragraph{Trajectories.}
Given a problem instance $x \sim \mathcal{D}$, executing controller $c$ yields a multi-turn trajectory

\(
\tau = \big( (m^{(1)}, y_{1,1:T_1}),\; \dots,\; (m^{(U_c)}, y_{U_c,1:T_{U_c}}) \big)
\),

where $U_c$ is the number of LLM calls (turns) under controller $c$, $m^{(u)} \in \mathcal{M}_c$ is the module active at turn $u$, and $y_{u,1:T_u}$ is the token sequence generated at that turn.
The same module may be invoked at multiple turns (e.g., the CoT module is called both before and after the critic in Self-Refine), and $U_c$ may vary across instances when the control flow includes conditional logic.

\paragraph{State and policy.}
At turn $u$, the model observes
\(
s_u = \big(x,\; \tau_{<u},\; \kappa^{(u)}\big)
\),
where $\tau_{<u}$ is the interaction history up to turn $u$ (as assembled by the control flow $f_c$) and $\kappa^{(u)}$ is the turn interface of the active module.
We train a single shared policy $\pi_\theta$ that generates tokens autoregressively conditioned on $s_u$, via
\(
\pi_\theta(y_{u,j} \mid s_{u,j})
\), where
\(
s_{u,j} = \big(x,\; \tau_{<u},\; y_{u,<j},\; \kappa^{(u)}\big)
\).

\paragraph{Reward.}
For reasoning tasks with verifiable answers, the reward is terminal and binary: $R(\tau) \in \{0,1\}$.
Intermediate turns receive no reward, except the format rewards which are applied at each turn (See Section~\ref{sec:exp_setup}).

\paragraph{Module vocabulary and compositionality.}
Across training controllers, we identify a shared \emph{module vocabulary} $\mathcal{M} = \{1, \dots, M\}$ by grouping turns whose modules share the same role instruction and output schema.
Table~\ref{tab:modules} lists an example of five module types that appear in our experiments.
The key observation is that different controllers compose these modules in different ways:
Chain-of-Thought uses only the \textsc{cot} module;
Step-back Abstraction~\citep{zheng2024take} sequences \textsc{abstractor}~$\to$~\textsc{cot};
Self-Refine~\citep{madaan2024selfrefine} sequences \textsc{cot}~$\to$~\textsc{critic}~$\to$~\textsc{cot}, with repeated interaction between \textsc{cot} and \textsc{critic} modules;
and LLM-Debate~\citep{du2023debate} interleaves multiple \textsc{debater} turns with a \textsc{judge}.
This compositional structure motivates our central hypothesis: if the shared policy learns to execute each module well, it can generalize to novel \emph{compositions} of the same modules, i.e., controller configurations not seen during training.

\paragraph{Training objective.}
Let $\mathcal{C}_{\mathrm{train}}$ denote the set of training controllers.
Since the test-time controller distribution is unavailable during training, we optimize
\begin{equation}
J_{\mathrm{train}}(\theta) = \frac{1}{|\mathcal{C}_{\mathrm{train}}|} \sum_{c \in \mathcal{C}_{\mathrm{train}}} \mathbb{E}_{x \sim \mathcal{D}}\, \mathbb{E}_{\tau \sim \pi_\theta(\cdot \mid x,c)} \big[ R(\tau) \big].
\label{eq:train_obj}
\end{equation}
This objective trains a single shared policy over a distribution of controller-induced MDPs, placing CALM closer to robust multi-task RL~\citep{teh2017distral, yu2020gradient} 

The compositional module structure further connects CALM to modular multi-task RL~\citep{andreas2017modular, devin2017learning}, though we use a single shared policy conditioned on local turn interfaces rather than explicit architectural modularity or learned routing.

\section{Controller Aware Language Model}
\label{sec:method}

We describe the CALM training framework in two parts.
Section~\ref{sec:turn_grpo} introduces the base multi-controller GRPO objective with turn-level aggregation.

Section~\ref{sec:strategies} presents several training strategies including module-level decomposition of this objective.

\subsection{Turn-level GRPO over controllers}
\label{sec:turn_grpo}

We adopt Group Relative Policy Optimization (GRPO)~\citep{shao2024deepseekmath} and extend it to the multi-controller setting using a turn-level aggregation following~\citet{wan2025rema}.
At each training step, a controller $c$ is sampled uniformly from $\mathcal{C}_{\mathrm{train}}$ and a problem $x$ is sampled from $\mathcal{D}$.
The control flow $f_c$ is then executed $G$ times to produce a group of trajectories
$\tau_i \sim \pi_\theta(\cdot \mid x, c)$, for $i = 1, \dots, G$,
each receiving a terminal reward $R_i \in \{0,1\}$.

\paragraph{Multi-controller CALM objective.}
The full CALM training objective averages the controller-level losses over the training controller distribution and adds a KL penalty to the reference policy:
\begin{equation}
\mathcal{L}_{\mathrm{CALM}}(\theta)
= \frac{1}{|\mathcal{C}_{\mathrm{train}}|}
  \sum_{c \in \mathcal{C}_{\mathrm{train}}}
  \mathcal{L}_c^{\mathrm{GRPO}}(\theta)
  \;+\; \beta\, \overline{\mathrm{KL}}
\label{eq:calm_obj}
\end{equation}
where $\overline{\mathrm{KL}} = \mathbb{E}\!\left[D_{\mathrm{KL}}\!\left(\pi_\theta \,\|\, \pi_{\mathrm{ref}}\right)\right]$ is the average KL divergence to the reference policy over all tokens in the batch.
In practice, each training step samples a single controller $c \sim \mathrm{Uniform}(\mathcal{C}_{\mathrm{train}})$ and a single problem instance $x \sim \mathcal{D}$, so the outer sum is approximated by a single-sample estimate. $\mathcal{L}_c^{\mathrm{GRPO}}(\theta)$ uses a same turn level objective similar to ~\cite{wan2025rema}.

\paragraph{Training instability in multi-turn RL.}
\label{sec:instability}
Multi-turn GRPO with terminal rewards presents known training stability challenges.
As the number of turns grows, the credit assignment problem intensifies: a single binary reward must be distributed across turns that may contribute very differently to the outcome.
In practice, we observe that training can become unstable after a moderate number of optimization steps, consistent with findings in prior multi-turn RL work~\citep{wan2025rema, wang2025ragen}.
Both single-controller and multi-controller configurations are affected.
We do not claim to solve this instability; rather, we design our training protocol and evaluation to account for it (Section~\ref{sec:experiments}).

\subsection{Multi-controller training variants}
\label{sec:module_decomp}
\label{sec:strategies}

We investigate four experimental variants of multi-controller training that differ in how gradient updates are distributed across module types. We start with the simplest starting point which trains all the modules together as they appear in a batch without any explicit decomposition. However, the compositional structure of controllers and the fact that different controllers share reusable module types enables a natural decomposition of the multi-controller loss by module. We describe the four training variants below with the experimental section showing systematic comparisons of these four training strategies.

\paragraph{Multi-controller mixing (\textsc{calm-mixed}).}
The first variant trains on multi-controller data \emph{without} module decomposition.
All turns from all controllers are placed in a single flat batch and optimized with the undifferentiated turn-level GRPO loss averaged across sampled controllers.
The effective weight of each module is proportional to the number of turns it contributes, so frequent modules like \textsc{cot} naturally receive more gradient signal.
This variant establishes a direct baseline for whether module-aware decomposition confers any benefit over straightforward multi-controller training. 

\paragraph{Module decomposed training strategies.}
We now describe different training strategies for module decomposed training.
Recall from Section~\ref{sec:formulation} that each turn $u$ in trajectory $i$ is associated with a module type $m^{(u)} \in \mathcal{M}$.
For module $m$, define the set of turns of that type within trajectory $i$:
\[
\mathcal{U}_i^{(m)} = \big\{u : m^{(u)} = m \big\}.
\]
The \emph{module-masked} GRPO loss isolates the contribution of module $m$ to the overall update:
\begin{equation}
\small
\begin{aligned}
\mathcal{L}_m^{\mathrm{GRPO}}(\theta)
&= -\frac{1}{G} \sum_{i=1}^{G}
\frac{1}{|\mathcal{U}_i^{(m)}| + \eta}
\sum_{u \in \mathcal{U}_i^{(m)}} \\
&\quad \min\Big(
r_{i,u}(\theta)\hat{A}_i^{(c)},\;
\mathrm{clip}\big(r_{i,u}(\theta), 1{-}\varepsilon, 1{+}\varepsilon\big)\hat{A}_i^{(c)}
\Big),
\end{aligned}
\label{eq:module_grpo}
\end{equation}where $\eta > 0$ avoids division by zero when trajectory $i$ contains no turn of type $m$.
The advantage $\hat{A}_i^{(c)}$ remains trajectory-level (Eq.~\ref{eq:adv}); the decomposition separates \emph{which turns contribute to the gradient}, but does not alter the credit assignment.

Given a batch that may contain trajectories from multiple modules, let $\mathcal{M}_{\mathrm{batch}}$ denote the set of module types present.
Thus, the objective for module decomposed training is:
\begin{equation}
\mathcal{L}_{\mathrm{CALM}_{Decomp}}(\theta)
= \sum_{m \in \mathcal{M}_{\mathrm{batch}}} w_m \, \mathcal{L}_m^{\mathrm{GRPO}}(\theta)
+ \beta\, \overline{\mathrm{KL}},
\label{eq:calm_decomp_obj}
\end{equation}
where $w_m$ are per-module weights and $\overline{\mathrm{KL}}$ is the average KL penalty over all tokens in the batch.
Different choices of $w_m$ yield different training strategies, which we describe next.

\paragraph{Uniform module weighting (\textsc{calm-uniform}).}
The second variant sets $w_m = 1/|\mathcal{M}_{\mathrm{batch}}|$ for all active modules, so each module type contributes equally to the gradient update independent of how many turns it occupies in the current batch.
This is the simplest form of module-aware decomposition and serves as the reference point for the decomposed variants.

\paragraph{Single-role-per-step training (\textsc{calm-single}).}
The third variant takes module isolation to its extreme: at each training step, after rolling out trajectories under the sampled controller, we randomly select one module type $m^* \sim \mathrm{Uniform}(\mathcal{M}_{\mathrm{batch}})$ and compute the loss only on turns of type $m^*$, discarding data from all other modules.
This eliminates cross-module gradient interaction within a step at the cost of discarding turn data.
It directly tests whether inter-module gradient interference is a significant obstacle to cross-controller generalization.

\paragraph{Adaptive gradient-norm weighting (\textsc{calm-adaptive}).}
The fourth variant investigates whether explicit gradient balancing across modules is beneficial.
Modules may induce very different gradient scales, causing a subset to dominate the shared-parameter update - a phenomenon analogous to task domination in multi-task learning~\citep{yu2020gradient, chen2018gradnorm}.
We maintain an exponential moving average of the gradient norm for each module:
\begin{equation}
\nu_m \leftarrow \rho\, \nu_m + (1 - \rho)\, \gamma_m,
\qquad
\gamma_m = \|\nabla_\theta \mathcal{L}_m^{\mathrm{GRPO}}(\theta)\|_2,
\label{eq:ema}
\end{equation}
and set module weights inversely proportional to the smoothed gradient norm:
\begin{equation}
\small
\begin{aligned}
w_m
&= \mathrm{clip}\Big(
\big(\tfrac{\bar{\nu}}{\nu_m + \delta}\big)^{\alpha},
\; w_{\min},\; w_{\max}
\Big), \\
\bar{\nu}
&= \frac{1}{|\mathcal{M}_{\mathrm{active}}|}
\sum_{m} \nu_m,
\end{aligned}
\label{eq:adaptive_weight}
\end{equation}
where $\delta > 0$ prevents division by zero, $\alpha$ controls the balancing strength, and clipping bounds prevent extreme weights.
Modules with persistently large gradients receive lower weight, and vice versa.
A warmup phase with uniform weights is used for the first $T_w$ steps to allow gradient norm estimates to stabilize.

As we discuss in Section~\ref{sec:experiments}, each variant has strengths in different evaluation regimes, and no single configuration dominates across all controller types.

\section{Related Work}
\label{app:related}
\paragraph{Inference-time computation and controllers.}
A growing body of work augments LLM reasoning by structuring inference across multiple calls or intermediate steps.
Foundational methods include Chain-of-Thought prompting~\citep{wei2022chainofthought}, Self-Consistency~\citep{wang2023selfconsistency}, and Tree-of-Thoughts~\citep{yao2023tree}.
More structured pipelines incorporate self-refinement~\citep{madaan2024selfrefine, shinn2023reflexion}, multi-agent debate~\citep{du2023debate}, and tool-augmented reasoning~\citep{schick2023toolformer}.
Recent work has begun automating the design and search of such inference-time procedures~\citep{saad2024archon, hu2025adas}.
CALM is complementary to this line of work: rather than optimizing the controller given a fixed model, we optimize the model given a \emph{distribution} of controllers.

\paragraph{Single-controller fine-tuning.}
Prior post-training work largely assumes a single fixed interaction pattern.
\citet{chow2025bon} study fine-tuning strategies specific to Best-of-N sampling; \citet{shinn2023reflexion} train agents for Reflexion-style verbal reinforcement; and RLHF and related methods~\citep{brown2020language} target single-turn or single-role interaction.
These approaches produce models that may over-specialize to one workflow and fail to transfer when familiar reasoning behaviors appear in a novel controller composition.
CALM addresses this directly by placing the training objective over a \emph{family} of controllers rather than a single fixed one.

\paragraph{Multi-task and robust reinforcement learning.}
Multi-task RL trains a shared policy over a distribution of tasks or
MDPs~\citep{caruana1997multitask, teh2017distral, yu2020gradient}.
Robust policy optimization seeks policies that remain effective under
distributional shift~\citep{tessler2019action, ghosh2021generalization}.
CALM is related to both: we train a shared policy over controller-induced MDPs and evaluate on held-out controller distributions.
However, the multi-controller setting introduces structure not present in standard multi-task RL.
Controllers are not independent tasks — they share reusable local reasoning modules while differing in composition and turn structure.
This heterogeneity produces module-level gradient imbalance that standard multi-task objectives do not address.
Our compositional multi-module RL objective is specifically designed for this setting, and to the best of our knowledge, no prior work has studied post-training over a distribution of inference-time controllers.

\paragraph{Modular and compositional reinforcement learning.}
A separate line of work studies policies composed from reusable sub-policies or skill primitives~\citep{devin2017learning, andreas2017modular, eysenbach2018diversity}.
These approaches often require explicit task decomposition, modular architectures, or learned routing between sub-policies.
CALM differs in that we use a single shared policy conditioned on local turn interfaces, without requiring architectural modularity or explicit module routing.
The compositional structure is instead reflected in the training objective through module-masked gradient updates and gradient-norm balancing.

\paragraph{Meta-learning for language models.}
Meta-learning approaches such as MAML~\citep{finn2017maml} and its variants optimize for fast adaptation to new tasks from limited experience.
While superficially related, the goal of meta-learning is distinct from CALM's objective: meta-learning targets rapid fine-tuning at test time, whereas CALM trains a single fixed policy that transfers to new controller compositions without any test-time gradient update.
This distinction matters in practice, since most deployment settings do not permit test-time optimization of model parameters.

\paragraph{Multi-turn and multi-agent RL for LLMs.}
Recent work has explored RL objectives for multi-turn LLM interaction~\citep{wan2025rema} and multi-agent
settings~\citep{wang2025ragen}.
CALM builds on the turn-level GRPO formulation of \citet{wan2025rema}, which aggregates token-level importance ratios at the turn level to handle variable-length multi-turn trajectories.
We extend this formulation to the multi-controller setting by introducing module-level loss decomposition and gradient-norm balancing to handle the heterogeneous turn structure induced by different controllers.

\section{Experiments}
\label{sec:experiments}

\begin{table*}[t]
\centering
\caption{Simplified view of controllers used in both experiments. \emph{Train}: seen during CALM training. \emph{Comp.}: novel compositions of known modules. \emph{Shift}: new module types introduced at test time. \checkmark{} marks the subset evaluated in Experiment~2.}
\label{tab:eval_controllers}
\scriptsize
\setlength{\tabcolsep}{5pt}
\begin{tabular}{@{}llllcc@{}}
\toprule
\textbf{ID} & \textbf{Controller} & \textbf{Modules} & \textbf{Tier} & \textbf{Exp.\ 1} & \textbf{Exp.\ 2} \\
\midrule
B1 & Chain-of-Thought     & \textsc{cot}                                                                   & Train & \checkmark & \checkmark \\
B2 & Self-Refine          & \textsc{cot}$\to$ (\textsc{critic}$\to$\textsc{cot}) $\times 2$                             & Train & \checkmark & \checkmark \\ 
B3 & LLM-Debate           & \textsc{debater}$\times$3$\to$\textsc{judge}                                  & Train & \checkmark &  \\
B4 & Step-back Abstr.     & \textsc{abstractor}$\to$\textsc{cot}                                          & Train & \checkmark & \\
B5 & Quality-Diversity    & \textsc{cot}$\times$3$\to$\textsc{judge}                                      & Train & \checkmark & \\
B6 & Dyn.\ Role Assign    & \textsc{router}$\to$\textsc{cot}                                              & Train & \checkmark & \\
\midrule
H1 & CoT Self-Consist.    & \textsc{cot}$\times$5$+$\,vote                                                & Comp. & \checkmark & \checkmark \\
H2 & Step-back + Refine   & \textsc{abst}$\to$\textsc{cot}$\to$\textsc{critic}$\to$\textsc{cot}           & Comp. & \checkmark & \checkmark \\
H3 & Debate + Refine      & \textsc{deb}$\times$3$\to$\textsc{judge}$\to$\textsc{critic}$\to$\textsc{cot} & Comp. & \checkmark & \\
\midrule
H4 & Dyn.\ Role-Playing   & \textsc{teacher}$\to$\textsc{student}$\to$\textsc{verifier}$\to\cdots$        & Shift & \checkmark & \checkmark \\
H5 & Reflective Verif.    & \textsc{num}$\to$\textsc{log}$\to$\textsc{heur}$\to$\textsc{verifier}         & Shift & \checkmark & \checkmark \\
H6 & Learning Agent       & \textsc{learn-agent}$\to$\textsc{feedback-agent}$\times$3                     & Shift & \checkmark & \\
\bottomrule
\end{tabular}
\end{table*}

We conduct two experiments to evaluate controller-aware post-training. Experiment 1 compares all four CALM variants against single-controller baselines across twelve evaluation controllers, evaluating in-distribution performance of multi-controller training.
Experiment~2 tests whether such training generalize to unseen controllers and out-of-distribution data.

\subsection{Experimental Setup}
\label{sec:exp_setup}

\textbf{Datasets and models.}
For both experiments, we train on 2{,}048 problems from the GSM8K~\citep{cobbe2021training} training split using \textsc{Llama-3.2-3B-Instruct}~\citep{grattafiori2024llama}. In Experiment~1, we evaluate the trained models on the GSM8K test set. For Experiment~{2}, we use MATH500~\citep{lightman2023let} and AMC23\footnote{ \url{https://huggingface.co/datasets/math-ai/amc23}} for testing on out-of-distribution data.
Reward is binary in both experiments: $R = 1$ if the final extracted answer is correct, $R = 0$ otherwise. In addition to the binary correctness reward, we add a lightweight format reward at every turn to encourage well-formed structured outputs. The format reward penalizes JSON parse errors $(-0.5)$, missing required output fields $(-0.2)$, and extraneous fields $(-0.1)$, and assigns zero otherwise. This ensures the shared policy maintains valid turn interfaces across all module types throughout training, without conflating formatting compliance with task correctness.

\textbf{Controllers.}
Experiment~1 uses six training controllers (B1--B6) and evaluates on the same set of controllers spanning three tiers, as listed in Table~\ref{tab:eval_controllers}.
\emph{Training controllers} (B1--B6) are seen during training; \emph{compositional controllers} (H1--H3) recombine known modules in novel arrangements; \emph{controller-shift controllers} (H4--H6) introduce entirely new module types. All controllers are derived from prior literature~\citep{hu2025adas}, with the exception of H2 and H3, which we design manually to highlight the compositional properties of controller structures.
Experiment~2 uses the same six controllers for CALM during training and evaluates on six controllers across the same three tiers, as listed in Table~\ref{tab:eval_controllers}. For Experiment~2, we choose CoT as the single controller candidate because it was the one of the strongest single-controller baselines in Experiment~1. The python code and description of controllers are available in Appendix~\ref{app:controller_code}

\textbf{Baselines.}
In Experiment~1, we compare the four CALM variants (Table~\ref{tab:results}) against single-controller baselines. The single controller baselines were each trained on one controller from \emph{Training Controllers}, using the same objective, directly exposing the cost of over-specialization.
In Experiment~2, we compare the single-controller CoT baseline (B1) against CALM variants to see if the learned behaviors are transferable.
All training hyperparameters are reported in Appendix~\ref{app:hyperparams}.

\subsection{Experiment 1: Performance Comparison between CALM vs. Single Controllers}
\label{sec:gsm8k_results}
\begin{table*}[t]
  \caption{Results across all evaluation controllers. Each row reports accuracy (\%) under the corresponding evaluation controller. Single-controller baselines are trained on one controller only. CALM variants are trained on all six training controllers. Base model represents applying the  controller zero-shot. Red-highlighted cells indicate matched controller pairs (e.g., B1-B1, \ldots, B6-B6). \textbf{Bold} indicates the best result per column.}
  \label{tab:results}
  \scriptsize
  \setlength{\tabcolsep}{4pt}
  \renewcommand{\arraystretch}{1.12}
  \definecolor{lred}{RGB}{255,220,220}
  \centering
  \begin{tabular}{lrrrrrrcrrrcrrrc}
    \toprule
     & \multicolumn{7}{c}{\textbf{Training}} & \multicolumn{4}{c}{\textbf{Compositional}} & \multicolumn{4}{c}{\textbf{Controller-Shift}} \\
    \cmidrule(lr){2-8} \cmidrule(lr){9-12} \cmidrule(lr){13-16}
    \textbf{Run} & \textbf{B1} & \textbf{B2} & \textbf{B3} & \textbf{B4} & \textbf{B5} & \textbf{B6} & \cellcolor{gray!12}\textbf{Avg} & \textbf{H1} & \textbf{H2} & \textbf{H3} & \cellcolor{gray!12}\textbf{Avg} & \textbf{H4} & \textbf{H5} & \textbf{H6} & \cellcolor{gray!12}\textbf{Avg} \\
    \midrule
    Base Model & 16.83 & 49.66 & 36.69 & 20.39 & 36.62 & 25.63 & \cellcolor{gray!12}30.97 & 16.91 & 55.57 & 57.70 & \cellcolor{gray!12}43.39 & 58.30 & 57.39 & 58.53 & \cellcolor{gray!12}58.07 \\
    CoT (B1) & \cellcolor{lred}75.74 & 57.70 & 74.22 & 76.27 & 65.43 & 76.88 & \cellcolor{gray!12}71.04 & 76.19 & 62.55 & 64.82 & \cellcolor{gray!12}67.85 & 74.30 & 75.44 & \textbf{74.98} & \cellcolor{gray!12}74.91 \\
    Self-Refine (B2) & 26.08 & \cellcolor{lred}58.23 & 34.95 & 36.62 & 31.01 & 30.93 & \cellcolor{gray!12}36.30 & 27.22 & 58.83 & 51.48 & \cellcolor{gray!12}45.84 & 53.30 & 53.75 & 53.45 & \cellcolor{gray!12}53.50 \\
    Debate (B3) & 40.86 & 58.98 & \cellcolor{lred}71.87 & 63.23 & 51.02 & 46.25 & \cellcolor{gray!12}55.37 & 41.09 & 65.58 & 60.80 & \cellcolor{gray!12}55.82 & 70.89 & 73.39 & 65.20 & \cellcolor{gray!12}69.83 \\
    Step-back (B4) & 74.75 & 57.39 & 74.75 & \cellcolor{lred}75.89 & \textbf{68.31} & 75.59 & \cellcolor{gray!12}71.11 & 74.60 & 65.88 & 64.44 & \cellcolor{gray!12}68.31 & 74.45 & 75.89 & 71.95 & \cellcolor{gray!12}74.10 \\
    Quality-Diversity (B5) & 38.89 & 59.36 & 76.27 & 65.28 & \cellcolor{lred}65.43 & 45.41 & \cellcolor{gray!12}58.44 & 39.88 & 61.41 & 64.29 & \cellcolor{gray!12}55.19 & 70.13 & 76.27 & 71.49 & \cellcolor{gray!12}72.63 \\
    RoleAssign (B6) & 74.37 & 43.97 & 72.18 & 73.39 & 55.04 & \cellcolor{lred}75.59 & \cellcolor{gray!12}65.76 & 75.21 & 49.89 & 52.99 & \cellcolor{gray!12}59.36 & 71.19 & 75.21 & 71.95 & \cellcolor{gray!12}72.78 \\
    \midrule
    \textsc{CALM-MIXED} & 76.50 & 66.11 & \textbf{77.48} & \textbf{76.57} & 65.28 & 78.77 & \cellcolor{gray!12}73.45 & 76.72 & \textbf{68.54} & 65.66 & \cellcolor{gray!12}70.31 & \textbf{76.57} & 76.72 & 73.77 & \cellcolor{gray!12}\textbf{75.69} \\
    \textsc{CALM-SINGLE} & 76.95 & 60.96 & 76.12 & 74.45 & 62.40 & 78.09 & \cellcolor{gray!12}71.49 & 77.41 & 60.65 & 59.29 & \cellcolor{gray!12}65.78 & 76.27 & 76.50 & 73.62 & \cellcolor{gray!12}75.46 \\
    \textsc{CALM-UNIFORM} & 77.10 & 63.84 & 77.03 & 75.97 & 67.10 & 78.17 & \cellcolor{gray!12}73.20 & 77.48 & 66.11 & 64.06 & \cellcolor{gray!12}69.22 & 72.63 & \textbf{77.10} & 70.81 & \cellcolor{gray!12}73.51 \\
    \textsc{CALM-ADAPTIVE} & \textbf{77.71} & \textbf{66.41} & 77.26 & 74.68 & 67.17 & \textbf{79.08} & \cellcolor{gray!12}\textbf{73.72} & \textbf{77.86} & 68.08 & \textbf{66.11} & \cellcolor{gray!12}\textbf{70.68} & 69.90 & 73.92 & 65.81 & \cellcolor{gray!12}69.88 \\
    \bottomrule
  \end{tabular}
\end{table*}

\begin{table*}[t]
\centering
\caption{Experiment 2 results on MATH500 and AMC 2023. \textbf{Bold} indicates the best result per column within each dataset.}
\label{tab:math_results}
\small
\setlength{\tabcolsep}{4pt}
\renewcommand{\arraystretch}{1.12}
\begin{tabular}{>{\raggedright\arraybackslash}m{1.8cm} l rrcrrcrrc}
\toprule
 & & \multicolumn{3}{c}{\textbf{Training}} & \multicolumn{3}{c}{\textbf{Compositional}} & \multicolumn{3}{c}{\textbf{Controller-Shift}} \\
\cmidrule(lr){3-5} \cmidrule(lr){6-8} \cmidrule(lr){9-11}
\textbf{Dataset} & \textbf{Run} & \textbf{B1} & \textbf{B2} & \cellcolor{gray!12}\textbf{Avg} & \textbf{H1} & \textbf{H2} & \cellcolor{gray!12}\textbf{Avg} & \textbf{H4} & \textbf{H5} & \cellcolor{gray!12}\textbf{Avg} \\
\midrule
\multirow{5}{=}{\textbf{MATH500}}
  & Chain-of-Thought    & \textbf{45.00} & 31.60 & \cellcolor{gray!12}38.30          & 33.40          & \textbf{34.20} & \cellcolor{gray!12}33.80          & \textbf{39.80} & 37.20          & \cellcolor{gray!12}38.50 \\
  & \textsc{CALM-MIXED} & 42.40 & 35.40 & \cellcolor{gray!12}\textbf{38.90} & \textbf{36.60} & 34.20 & \cellcolor{gray!12}\textbf{35.40} & 39.00 & \textbf{41.20} & \cellcolor{gray!12}\textbf{40.10} \\
  & \textsc{CALM-SINGLE} & 41.20 & 35.40 & \cellcolor{gray!12}38.30 & 33.60 & 33.40 & \cellcolor{gray!12}33.50 & 37.60 & 38.40 & \cellcolor{gray!12}38.00 \\
   & \textsc{CALM-UNIFORM} & 40.60 & 34.80 & \cellcolor{gray!12}37.70 & 34.20 & 35.80 & \cellcolor{gray!12}35.00 & 39.40 & 36.60 & \cellcolor{gray!12}38.00 \\
   & \textsc{CALM-ADAPTIVE} & 42.00 & \textbf{35.80} & \cellcolor{gray!12}\textbf{38.90} & 33.60 & \textbf{36.60} & \cellcolor{gray!12}35.10 & \textbf{41.20} & 37.20 & \cellcolor{gray!12}39.20 \\
\midrule
\multirow{5}{=}{\textbf{AMC 2023}}
  & Chain-of-Thought    & 22.50          & 10.00          & \cellcolor{gray!12}16.25          & 15.00          & 22.50          & \cellcolor{gray!12}18.75          & 27.50          & \textbf{17.50} & \cellcolor{gray!12}22.50 \\
  & \textsc{CALM-MIXED} & \textbf{25.00} & 15.00 & \cellcolor{gray!12}\textbf{20.00} & \textbf{27.50} & \textbf{25.00} & \cellcolor{gray!12}\textbf{26.25} & \textbf{32.50} & 15.00 & \cellcolor{gray!12}23.75 \\
   & \textsc{CALM-SINGLE} & 22.50 & 15.00 & \cellcolor{gray!12}18.75 & 15.00 & 15.00 & \cellcolor{gray!12}15.00 & 22.50 & 17.50 & \cellcolor{gray!12}20.00 \\
   & \textsc{CALM-UNIFORM} & 15.00 & 15.00 & \cellcolor{gray!12}15.00 & 7.50 & 22.50 & \cellcolor{gray!12}15.00 & 25.00 & \textbf{27.50} & \cellcolor{gray!12}\textbf{26.25} \\
   & \textsc{CALM-ADAPTIVE} & 20.00 & \textbf{17.50} & \cellcolor{gray!12}18.75 & 15.00 & 12.50 & \cellcolor{gray!12}13.75 & 25.00 & 25.00 & \cellcolor{gray!12}25.00 \\
\bottomrule
\end{tabular}
\end{table*}
Table~\ref{tab:results} reports accuracy for single-controller baselines and all four CALM variants. All variants are trained for 3 epochs ($192$ steps). For Self-Refine (B2), we report results at the best validation checkpoint (step $156$), as training becomes unstable beyond that point. We refer to this as the known phenomenon of multi-turn training instability (See Section~\ref{sec:instability}).

\paragraph{Single-controller baselines over-specialize.}
Baselines trained on a single workflow perform well under their own controller but degrade sharply elsewhere.
Debate (B3) achieves $71.87\%$ on its own controller but drops to $40.86\%$ on Chain-of-Thought (B1); Quality-Diversity (B5) similarly collapses to $38.89\%$ on B1. Self-Refine achieves the lowest on average on training controllers ($36.30\%$), although it achieves ($58.23\%)$ in its own controller.
In contrast, CoT (B1, avg.\ $71.04\%$) and Step-back (B4, avg.\ $71.11\%$) are the strongest single-controller baselines because the CoT module they rely on appears in every evaluation controller. Furthermore, RoleAssign (B6) performs ($43.97\%$) even worse in B2 than the base model ($49.66\%$) with no additional RL training which indicates overfitting to a particular controller type even though it performs well in its own controller ($75.59\%$).

\paragraph{All CALM variants improve over single-controller baselines.}
All four CALM variants outperform the best single-controller baseline (Step-back, $71.11\%$) on the training-controller average, with averages ranging from $71.49\%$ (\textsc{calm-single}) to $73.72\%$ (\textsc{calm-adaptive}).
The gain is largest on Self-Refine (B2), where single-controller baselines span $43.97\%-59.36\%$ but CALM variants reach $60.96\%-66.41\%$, confirming that multi-controller training learns critic-style turns more robustly than any individual controller.
On compositional controllers, \textsc{calm-adaptive} ($70.68\%$) and \textsc{calm-mixed} ($70.31\%$) lead, both above the best single-controller baseline (Step-back, $68.31\%$); \textsc{calm-single} ($65.78\%$) lags, suggesting that discarding cross-module turn data hurts performance on longer interleaved compositions (H2, H3).

\paragraph{No variant dominates under controller shift.}
\textsc{calm-mixed} ($75.69\%$) and \textsc{calm-single} ($75.46\%$) are the strongest under controller shift, both exceeding the CoT baseline ($74.91\%$) by a small margin.
However, \textsc{calm-adaptive} drops to $69.98\%$ which is below the CoT baseline despite leading on compositional controllers, revealing that adaptive gradient-norm balancing can over-fit to the training module distribution and hurt robustness under broader structural change.
Overall, no single variant dominates across all three tiers: \textsc{calm-adaptive} is best when the test controllers recombine familiar modules; \textsc{calm-mixed} is most robust when the controller distribution may shift substantially.

\subsection{Experiment 2: Performance Comparison on Out-of-distribution Dataset}
For this experiment, we take the trained models from Experiment 1 and test it with out-of-distribution math datasets. Table~\ref{tab:math_results} reports results on MATH500 and AMC 2023, which out-of-distribution evaluations  for the MATH Level 3--5 training data.
On MATH500, \textsc{calm-mixed} and \textsc{calm-adaptive} jointly achieve the highest training average ($38.90\%$), and \textsc{calm-mixed} leads on both compositional ($35.40\%$) and controller-shift ($40.10\%$) tiers, while CoT remains competitive on individual controllers such as B1 ($45.00\%$) and H4 ($39.80\%$).
The harder AMC 2023 benchmark reveals a clearer separation: \textsc{calm-mixed} leads across all three tiers with averages of $20.00\%, 26.25\%$, and $23.75\%$, substantially outperforming the CoT baseline ($16.25\%, 18.75\%, 22.50\%$).
\textsc{calm-uniform} achieves the strongest controller-shift average on AMC 2023 ($26.25\%$), driven by a strong H5 result ($27.50\%$), while \textsc{calm-adaptive} underperforms on compositional controllers ($13.75\%$).
Overall, multi-controller training generalizes to both harder in-distribution problems and out-of-distribution competition benchmarks, with \textsc{calm-mixed} providing the most consistent gains across tiers and datasets.

\section{Conclusion}
We present CALM as a framework to address a fundamental mismatch in how LLMs are post-trained. As the discovery of inference-time techniques continue to advance, the question of how a base model should be trained to cooperate effectively with a diverse and evolving family of controllers remains largely open.
We present four different variations of multi-controller post training. Our experiments on GSM8K show that multi-controller training consistently outperforms single-controller baselines, which degrade sharply when evaluated outside their native workflow, and that the benefit extends to held-out controller compositions and broader controller shifts involving new module types. No single module-weighting variant dominates across all settings, suggesting that the choice of training strategy should be informed by the anticipated test-time controller distribution. More broadly, CALM points to a productive direction for future work: bridging the gap between two complementary lines of research: the automatic discovery of inference-time techniques, and the post-training of base models to use them effectively.

\section{Impact Statement}
This paper presents work whose goal is to advance the field of Compositional Learning for LLM Agents. There are many potential societal consequences of our work, none of which we feel must be specifically highlighted here.

\bibliography{example_paper}
\bibliographystyle{icml2026}

\clearpage
\onecolumn
\appendix
\renewcommand{\thesection}{\Alph{section}}

\section*{Appendix Table of Contents}

\makeatletter
\noindent\textbf{A\quad Multi-controller Training Objective}
\dotfill \pageref{app:turn_grpo}\par
\noindent\textbf{B\quad Module Vocabulary and Controller Pseudocode}
\dotfill \pageref{app:controllers}\par
\noindent\textbf{C\quad Hyperparameters and Training Details}
\dotfill \pageref{app:hyperparams}\par
\noindent\textbf{D\quad Training Plots}
\dotfill \pageref{app:training_plots}\par
\noindent\textbf{E\quad Qualitative Examples}
\dotfill \pageref{app:qualitative}\par
\makeatother

\clearpage
\section{Multi-controller Training Objective}
\label{app:turn_grpo}

We adopt Group Relative Policy Optimization (GRPO)~\citep{shao2024deepseekmath} and extend it to the multi-controller setting using a turn-level aggregation following~\citet{wan2025rema}.
At each training step, a controller $c$ is sampled uniformly from $\mathcal{C}_{\mathrm{train}}$ and a problem $x$ is sampled from $\mathcal{D}$.
The control flow $f_c$ is then executed $G$ times to produce a group of trajectories
\[
\tau_i \sim \pi_\theta(\cdot \mid x, c), \qquad i = 1, \dots, G,
\]
each receiving a terminal reward $R_i \in \{0, 1\}$.

\paragraph{Group-normalized advantages.}
Since reward is terminal, all turns within a trajectory share the same advantage.
We normalize within the group of $G$ trajectories sampled for the same controller and input:
\begin{equation}
\hat{A}_i^{(c)} = \frac{R_i - \mu_c}{\sigma_c + \epsilon},
\qquad
\mu_c = \frac{1}{G} \sum_{k=1}^{G} R_k,
\qquad
\sigma_c = \sqrt{\frac{1}{G} \sum_{k=1}^{G} (R_k - \mu_c)^2}.
\label{eq:adv}
\end{equation}
Note that the advantage $\hat{A}_i^{(c)}$ is identical for every turn $u$ within trajectory $i$: the decomposition does not perform turn-level credit assignment.

\paragraph{Turn-level ratio aggregation.}
For token $j$ at turn $u$ of trajectory $i$, the importance ratio is
\begin{equation}
r_{i,u,j}(\theta) = \frac{\pi_\theta\!\left(y_{i,u,j} \mid s_{i,u,j}\right)}{\pi_{\theta_{\mathrm{old}}}\!\left(y_{i,u,j} \mid s_{i,u,j}\right)},
\label{eq:token_ratio}
\end{equation}
where $s_{i,u,j} = (x, \tau_{i,<u}, y_{i,u,<j}, \kappa^{(u)})$ is the state at token $j$ of turn $u$ (see Section~\ref{sec:formulation}). 
Following~\citet{wan2025rema}, we aggregate ratios at the turn level to avoid length bias across turns with different token counts:
\begin{equation}
r_{i,u}(\theta) = \frac{1}{T_{i,u}} \sum_{j=1}^{T_{i,u}} r_{i,u,j}(\theta),
\label{eq:turn_ratio}
\end{equation}
where $T_{i,u}$ is the number of tokens generated at turn $u$ of trajectory $i$.

\paragraph{Controller-level GRPO loss.}
For a single sampled controller $c$, the turn-level GRPO loss over the group of $G$ trajectories is:
\begin{equation}
\mathcal{L}_c^{\mathrm{GRPO}}(\theta)
= -\frac{1}{G} \sum_{i=1}^{G} \frac{1}{U_i} \sum_{u=1}^{U_i} \frac{1}{T_{i,u}} \sum_{j=1}^{T_{i,u}}
\min\!\Big(
  r_{i,u}(\theta)\,\hat{A}_i^{(c)},\;
  \mathrm{clip}\big(r_{i,u}(\theta),\, 1{-}\varepsilon,\, 1{+}\varepsilon\big)\,\hat{A}_i^{(c)}
\Big),
\label{eq:controller_grpo_app}
\end{equation}
where $U_i$ is the number of turns in trajectory $i$ and $\varepsilon$ is the clipping parameter.
The normalization by $1/U_i$ is important in the multi-controller setting because controllers differ substantially in the number of turns (e.g., 1 turn for Chain-of-Thought vs.\ 7 turns for LLM-Debate): without it, longer controllers would dominate the loss.

\paragraph{Multi-controller CALM objective.}
The full CALM training objective averages the controller-level losses over the training controller distribution and adds a KL penalty to the reference policy:
\begin{equation}
\boxed{
\mathcal{L}_{\mathrm{CALM}}(\theta)
= \frac{1}{|\mathcal{C}_{\mathrm{train}}|}
  \sum_{c \in \mathcal{C}_{\mathrm{train}}}
  \mathcal{L}_c^{\mathrm{GRPO}}(\theta)
  \;+\; \beta\, \overline{\mathrm{KL}}
}
\label{eq:calm_obj_app}
\end{equation}
where $\overline{\mathrm{KL}} = \mathbb{E}\!\left[D_{\mathrm{KL}}\!\left(\pi_\theta \,\|\, \pi_{\mathrm{ref}}\right)\right]$ is the average KL divergence to the reference policy over all tokens in the batch.
In practice, each training step samples a single controller $c \sim \mathrm{Uniform}(\mathcal{C}_{\mathrm{train}})$ and a single problem $x \sim \mathcal{D}$, so the outer sum is approximated by a single-sample estimate.

Expanding Eq.~\ref{eq:calm_obj_app} fully with the turn-level ratio (Eq.~\ref{eq:turn_ratio}) and advantage (Eq.~\ref{eq:adv}), the complete objective is:
\begin{align}
& \mathcal{L}_{\mathrm{CALM}}(\theta)
= -\frac{1}{|\mathcal{C}_{\mathrm{train}}|}
\sum_{c \in \mathcal{C}_{\mathrm{train}}}
\frac{1}{G} \sum_{i=1}^{G}
\frac{1}{U_i} \sum_{u=1}^{U_i} \frac{1}{T_{i,u}} \sum_{j=1}^{T_{i,u}}\nonumber \\
&\quad
\min\!\bigg(
  \underbrace{\frac{1}{T_{i,u}} \sum_{j=1}^{T_{i,u}} r_{i,u,j}(\theta)}_{\text{turn-level ratio } r_{i,u}(\theta)}
  \;\hat{A}_i^{(c)},\;\;
  \mathrm{clip}\!\Big(\frac{1}{T_{i,u}} \sum_{j=1}^{T_{i,u}} r_{i,u,j}(\theta),\; 1{-}\varepsilon,\; 1{+}\varepsilon\Big)
  \;\hat{A}_i^{(c)}
\bigg)
\;+\; \beta\,\overline{\mathrm{KL}},
\label{eq:calm_expanded}
\end{align}
where the four levels of summation correspond to: (1)~controllers $c$ sampled from the training family, (2)~trajectories $i$ within a group of $G$ rollouts, (3)~turns $u$ within each trajectory, normalized by $1/U_i$, and (4)~tokens $j$ within each turn, aggregated into the turn-level ratio $r_{i,u}(\theta)$.
This objective reduces to single-controller turn-level GRPO (Eq.~4 of \citet{wan2025rema}) when $|\mathcal{C}_{\mathrm{train}}| = 1$, and to standard single-turn GRPO~\citep{shao2024deepseekmath} when additionally $U_i = 1$ for all trajectories.

\section{Module Vocabulary and Controller Pseudocode}
\label{app:controllers}

\subsection{Module vocabulary}
\label{app:modules}

Table~\ref{tab:modules} lists the full module vocabulary used in our experiments.
Each module is defined by a role instruction (system prompt) and an output schema.
The shared module vocabulary is the key structure that makes cross-controller generalization possible: a \textsc{critic} turn presents the same interface regardless of whether it appears in Self-Refine or in a held-out compositional controller. However, the modules could have distinct prompt with optionally associated roles (See Section~\ref{app:controller_code}). 

\begin{table}[h]
\centering
\caption{Module vocabulary used across all training and evaluation controllers.
Each module is identified by its role label and output schema; these are the only controller-specific inputs the shared policy observes at each turn.}
\label{tab:modules}
\small
\begin{tabular}{@{}lll@{}}
\toprule
\textbf{Module} & \textbf{Role description} & \textbf{Output schema} \\
\midrule
\textsc{cot}        & Chain-of-thought reasoning agent   & \texttt{[thinking, answer]} \\
\textsc{critic}     & Critique and feedback agent        & \texttt{[feedback, correct]} \\
\textsc{abstractor} & Step-back abstraction agent        & \texttt{[thinking, principle]} \\
\textsc{debater}    & Multi-agent debate participant     & \texttt{[thinking, answer]} \\
\textsc{judge}      & Aggregation and verdict agent      & \texttt{[thinking, answer]} \\
\textsc{router}     & Dynamic role assignment agent      & \texttt{[choice]} \\
\bottomrule
\end{tabular}
\end{table}

\subsection{Controller implementations}
\label{app:controller_code}

We implement each controller as a Python callable that takes a problem instance and a policy function and returns a trajectory.
The controller specifies the module sequence, routes outputs between turns, and handles any non-LLM logic (branching, iteration, voting). Specifically, each controller is implemented as a Python method using \texttt{LLMAgentBase}, which takes a list of output fields, an agent name, and optional role/temperature arguments. For evaluation, we choose greedy sampling with temperature$=0$.
Agents are called with a list of context \texttt{Info} objects and a system instruction.
The six training controllers are shown below:
 
\begin{lstlisting}[caption={Chain-of-Thought (B1)}]
def Chain_of_Thought(self, taskInfo):
    # Instruction for the Chain-of-Thought (CoT) approach
    # It is a important practice that allow the LLM to think step by step before solving the task.
    cot_instruction = "Please think step by step and then solve the task."

    # Instantiate a new LLM agent specifically for CoT
    # To alow LLM thinking before answering, we need to set the an addtional output field 'thinking'.
    cot_agent = LLMAgentBase(['thinking', 'answer'], 'Chain-of-Thought Agent')

    # Prepare the inputs for the CoT agent
    # The input should be a list of Info, and the first one is often the taskInfo
    cot_agent_inputs = [taskInfo]

    # Get the response from the CoT agent
    thinking, answer = cot_agent(cot_agent_inputs, cot_instruction)

    # Return only the final answer
    return answer
\end{lstlisting}
 
\begin{lstlisting}[caption={Self-Refine (B2)}]
def Self_Refine_Reflexion(self, taskInfo):
    # Instruction for initial reasoning
    cot_initial_instruction = "Please think step by step and then solve the task."

    # Instruction for reflecting on previous attempts and feedback to improve
    cot_reflect_instruction = "Given previous attempts and feedback, carefully consider where you could go wrong in your latest attempt. Using insights from previous attempts, try to solve the task better."
    cot_agent = LLMAgentBase(['thinking', 'answer'], 'Chain-of-Thought Agent')

    # Instruction for providing feedback and correcting the answer
    critic_instruction = "Please review the answer above and criticize on where might be wrong. If you are absolutely sure it is correct, output 'True' in 'correct'."
    critic_agent = LLMAgentBase(['feedback', 'correct'], 'Critic Agent')
    
    N_max = 2 # Maximum number of attempts

    # Initial attempt
    cot_inputs = [taskInfo]
    thinking, answer = cot_agent(cot_inputs, cot_initial_instruction, 0)

    for i in range(N_max):
        # Get feedback and correct status from the critic
        feedback, correct = critic_agent([taskInfo, thinking, answer], critic_instruction, i)
        if correct.content == 'True':
            break
            
        # Add feedback to the inputs for the next iteration
        cot_inputs.extend([thinking, answer, feedback])

        # Reflect on previous attemps and refine the answer
        thinking, answer = cot_agent(cot_inputs, cot_reflect_instruction, i + 1)
    return answer
\end{lstlisting}
 
\begin{lstlisting}[caption={LLM-Debate (B3)}]
def LLM_Debate(self, taskInfo):
    # Instruction for initial reasoning
    debate_initial_instruction = "Please think step by step and then solve the task."

    # Instruction for debating and updating the solution based on other agents' solutions
    debate_instruction = "Given solutions to the problem from other agents, consider their opinions as additional advice. Please think carefully and provide an updated answer."
    
    # Initialize debate agents with different roles and a moderate temperature for varied reasoning
    debate_agents = [LLMAgentBase(['thinking', 'answer'], 'Debate Agent', temperature=0.8, role=role) for role in ['Math Professor', 'Grade School Teacher', 'Math Enthusiast']]

    # Instruction for final decision-making based on all debates and solutions
    final_decision_instruction = "Given all the above thinking and answers, reason over them carefully and provide a final answer."
    final_decision_agent = LLMAgentBase(['thinking', 'answer'], 'Final Decision Agent', temperature=0.1)

    max_round = 2 # Maximum number of debate rounds
    all_thinking = [[] for _ in range(max_round)]
    all_answer = [[] for _ in range(max_round)]

    # Perform debate rounds
    for r in range(max_round):
        for i in range(len(debate_agents)):
            if r == 0:
                thinking, answer = debate_agents[i]([taskInfo], debate_initial_instruction)
            else:
                input_infos = [taskInfo] + [all_thinking[r-1][i]] + all_thinking[r-1][:i] + all_thinking[r-1][i+1:]
                thinking, answer = debate_agents[i](input_infos, debate_instruction)
            all_thinking[r].append(thinking)
            all_answer[r].append(answer)
    
    # Make the final decision based on all debate results and solutions
    thinking, answer = final_decision_agent([taskInfo] + all_thinking[max_round-1] + all_answer[max_round-1], final_decision_instruction)
    return answer

\end{lstlisting}
 
\begin{lstlisting}[caption={Step-back Abstraction (B4)}]
def Step_back_Abstraction(self, taskInfo):
        # Instruction for understanding the principles involved in the task
        principle_instruction = "What are the physics, chemistry or biology principles and concepts involved in solving this task? First think step by step. Then list all involved principles and explain them."
        
        # Instruction for solving the task based on the principles
        cot_instruction = "Given the question and the invovled principle behind the question, think step by step and then solve the task."
        
        # Instantiate LLM agents
        principle_agent = LLMAgentBase(['thinking', 'principle'], 'Principle Agent')
        cot_agent = LLMAgentBase(['thinking', 'answer'], 'Chain-of-Thought Agent')
        
        # Get the principles involved in the task
        thinking, principle = principle_agent([taskInfo], principle_instruction)

        # Use the principles to solve the task
        thinking, answer = cot_agent([taskInfo, thinking, principle], cot_instruction)
        return answer
\end{lstlisting}
 
\begin{lstlisting}[caption={Quality-Diversity (B5)}]
def Quality_Diversity(self, taskInfo):
    # Instruction for initial reasoning
    cot_initial_instruction = "Please think step by step and then solve the task."

    # Instruction for giving diverse answers
    qd_instruction = "Given previous attempts, try to come up with another interesting way to solve the task."
    cot_agent = LLMAgentBase(['thinking', 'answer'], 'Chain-of-Thought Agent')

    # Instruction for final decision-making based on collected reasoning and answers
    final_decision_instruction = "Given all the above solutions, reason over them carefully and provide a final answer."
    final_decision_agent = LLMAgentBase(['thinking', 'answer'], 'Final Decision Agent', temperature=0.1)
    
    N_max = 2 # Maximum number of attempts

    # Initial attempt
    cot_inputs = [taskInfo]
    possible_answers = []
    thinking, answer = cot_agent(cot_inputs, cot_initial_instruction, 0)

    # Add the answer to the list of possible answers
    possible_answers.extend([thinking, answer])

    for i in range(N_max):
        # Reflect on previous attempts and generate another interesting answer
        cot_inputs.extend([thinking, answer])

        # Generate another interesting answer
        thinking, answer = cot_agent(cot_inputs, qd_instruction, i + 1)
        possible_answers.extend([thinking, answer])

    # Make the final decision based on all generated answers
    thinking, answer = final_decision_agent([taskInfo] + possible_answers, final_decision_instruction)
    return answer
\end{lstlisting}
 
\begin{lstlisting}[caption={Dynamic Role Assignment (B6)}]
def Dynamic_Assignment_of_Roles(self, taskInfo):
        # Instruction for step-by-step reasoning
        cot_instruction = "Please think step by step and then solve the task."
        expert_agents = [LLMAgentBase(['thinking', 'answer'], 'Expert Agent', role=role) for role in ['Math Professor', 'Grade School Teacher', 'Math Enthusiast', 'Helpful Assistant']]

        # Instruction for routing the task to the appropriate expert
        routing_instruction = "Given the task, please choose an Expert to answer the question. Choose from: Math Professor, Grade School Teacher, Math Enthusiast."
        routing_agent = LLMAgentBase(['choice'], 'Routing agent')

        # Get the choice of expert to route the task
        choice = routing_agent([taskInfo], routing_instruction)[0]

        if 'professor' in choice.content.lower():
            expert_id = 0
        elif 'teacher' in choice.content.lower():
            expert_id = 1
        elif 'enthusiast' in choice.content.lower():
            expert_id = 2
        else:
            expert_id = 3 # Default to helpful assistant

        thinking, answer = expert_agents[expert_id]([taskInfo], cot_instruction)
        return answer
\end{lstlisting}

Evaluation controllers (H1--H3) are constructed by composing these modules in novel arrangements as described in Table~\ref{tab:eval_controllers}.
The controller-shift controllers (H4--H6) additionally introduce module types not present in the training vocabulary (\textsc{teacher}, \textsc{student}, \textsc{verifier}, \textsc{learning-agent}, \textsc{feedback-agent}, and specialized verifier subtypes).

\begin{lstlisting}[caption={Self-Consistency with Chain-of-Thought (H1)}]
def Self_Consistency_with_Chain_of_Thought(self, taskInfo):
    # Instruction for step-by-step reasoning
    cot_instruction = "Please think step by step and then solve the task."
    N = 5 # Number of CoT agents

    # Initialize multiple CoT agents with a higher temperature for varied reasoning
    cot_agents = [LLMAgentBase(['thinking', 'answer'], 'Chain-of-Thought Agent', temperature=0.8) for _ in range(N)]

    # Majority voting function to select the most common answer
    from collections import Counter
    def majority_voting(answers):
        return Counter(answers).most_common(1)[0][0]
    
    possible_answers = []
    for i in range(N):
        thinking, answer = cot_agents[i]([taskInfo], cot_instruction)
        possible_answers.append(answer.content)

    # Ensembling the answers from multiple CoT agents
    answer = majority_voting(possible_answers)
    return answer  
\end{lstlisting}

\begin{lstlisting}[caption={Step-back with Refine (H2)}]
def Step_back_with_Refine(self, taskInfo):
    # === Module 1: ABSTRACTOR ===
    principle_instruction = "What are the physics, chemistry or biology principles and concepts involved in solving this task? First think step by step. Then list all involved principles and explain them."
    principle_agent = LLMAgentBase(['thinking', 'principle'], 'Principle Agent')

    # === Module 2: COT  ===
    cot_instruction = "Given the question and the invovled principle behind the question, think step by step and then solve the task."
    cot_agent = LLMAgentBase(['thinking', 'answer'], 'Chain-of-Thought Agent')

    # === Module 3: CRITIC ===
    critic_instruction = "Please review the answer above and criticize on where might be wrong. If you are absolutely sure it is correct, output 'True' in 'correct'."
    critic_agent = LLMAgentBase(['feedback', 'correct'], 'Critic Agent')

    # === Module 4: COT-reflect  ===
    cot_reflect_instruction = "Given previous attempts and feedback, carefully consider where you could go wrong in your latest attempt. Using insights from previous attempts, try to solve the task better."

    # Turn 1: Extract principles (ABSTRACTOR)
    thinking_p, principle = principle_agent([taskInfo], principle_instruction)

    # Turn 2: Solve using principles (COT)
    thinking, answer = cot_agent([taskInfo, thinking_p, principle], cot_instruction)

    # Turn 3: Critique the answer (CRITIC)
    feedback, correct = critic_agent([taskInfo, thinking, answer], critic_instruction)

    # Turn 4: Revise based on feedback (COT-reflect)
    thinking, answer = cot_agent([taskInfo, thinking_p, principle, thinking, answer, feedback], cot_reflect_instruction, 1)

    return answer
\end{lstlisting}

\begin{lstlisting}[caption={Debate + Refine (H3)}]
    def H5__Debate___Refine(self, taskInfo):
    # === Modules 1-3: DEBATERS ===
    debate_initial_instruction = "Please think step by step and then solve the task."
    debate_agents = [LLMAgentBase(['thinking', 'answer'], 'Debate Agent', temperature=0.8, role=role) for role in ['Math Professor', 'Grade School Teacher', 'Math Enthusiast']]

    # === Module 4: JUDGE ===
    final_decision_instruction = "Given all the above thinking and answers, reason over them carefully and provide a final answer."
    final_decision_agent = LLMAgentBase(['thinking', 'answer'], 'Final Decision Agent', temperature=0.1)

    # === Module 5: CRITIC  ===
    critic_instruction = "Please review the answer above and criticize on where might be wrong. If you are absolutely sure it is correct, output 'True' in 'correct'."
    critic_agent = LLMAgentBase(['feedback', 'correct'], 'Critic Agent')

    # === Module 6: COT-reflect ===
    cot_reflect_instruction = "Given previous attempts and feedback, carefully consider where you could go wrong in your latest attempt. Using insights from previous attempts, try to solve the task better."
    cot_agent = LLMAgentBase(['thinking', 'answer'], 'Chain-of-Thought Agent')

    # Turns 1-3: One round of debate (DEBATERS)
    all_thinking = []
    all_answer = []
    for i in range(3):
        thinking, answer = debate_agents[i]([taskInfo], debate_initial_instruction)
        all_thinking.append(thinking)
        all_answer.append(answer)

    # Turn 4: Judge decides (JUDGE)
    thinking_j, answer_j = final_decision_agent([taskInfo] + all_thinking + all_answer, final_decision_instruction)

    # Turn 5: Critic reviews judge's answer (CRITIC)
    feedback, correct = critic_agent([taskInfo, thinking_j, answer_j], critic_instruction)

    # Turn 6: Revise based on feedback (COT-reflect)
    thinking, answer = cot_agent([taskInfo, thinking_j, answer_j, feedback], cot_reflect_instruction, 1)

    return answer
\end{lstlisting}

\begin{lstlisting}[caption={Dynamic Role-Playing Architecture (H4)}]
def Dynamic_Role_Playing_Architecture(self, taskInfo):
    # Instruction for the Teacher agent to explain the task and provide feedback
    teacher_instruction = "Please explain the task in detail and provide any necessary hints or steps to solve it."

    # Instruction for the Student agent to solve the task based on the Teacher's explanation
    student_instruction = "Using the Teacher's explanation and feedback, please solve the task step by step."

    # Instruction for the Verifier agent to review the Student's answer and provide a final decision
    verifier_instruction = "Please review the Student's answer, verify its correctness, and provide a final decision. Also, give feedback on the Teacher's explanation and the Student's approach."

    # Initialize the Teacher, Student, and Verifier agents
    teacher_agent = LLMAgentBase(['explanation'], 'Teacher Agent')
    student_agent = LLMAgentBase(['thinking', 'answer'], 'Student Agent')
    verifier_agent = LLMAgentBase(['final_verdict', 'teacher_feedback', 'student_feedback'], 'Verifier Agent')

    # Teacher agent provides explanation and hints
    teacher_output = teacher_agent([taskInfo], teacher_instruction)
    explanation = teacher_output[0]

    # Student agent attempts to solve the task based on Teacher's explanation
    student_output = student_agent([taskInfo, explanation], student_instruction)
    thinking, answer = student_output

    # Verifier agent reviews the Student's answer and provides feedback
    verifier_output = verifier_agent([taskInfo, thinking, answer], verifier_instruction)
    final_verdict, teacher_feedback, student_feedback = verifier_output

    # Teacher agent refines the explanation based on Verifier's feedback
    refined_teacher_output = teacher_agent([taskInfo, teacher_feedback], teacher_instruction)
    refined_explanation = refined_teacher_output[0]

    # Student agent refines the solution based on refined Teacher's explanation and Verifier's feedback
    refined_student_output = student_agent([taskInfo, refined_explanation, student_feedback], student_instruction)
    refined_thinking, refined_answer = refined_student_output

    # Return the final answer
    return refined_answer
\end{lstlisting}
    
\begin{lstlisting}[caption={Role-based Reflective Peer Review (H5)}]
    
def Role_based_Reflective_Peer_Review(self, taskInfo):
    # Instructions for different reasoning styles
    numerical_instruction = 'Please solve the task using numerical calculations.'
    logical_instruction = 'Please solve the task using logical reasoning.'
    heuristic_instruction = 'Please solve the task using heuristic reasoning.'

    # Instruction for reflective verification
    reflective_verification_instruction = 'Please review the solutions provided by other agents, critique where they might be wrong, and suggest improvements.'

    # Initialize agents with different roles and approaches
    numerical_agent = LLMAgentBase(['thinking', 'answer'], 'Numerical Solver', temperature=0.3, role='numerical solver')
    logical_agent = LLMAgentBase(['thinking', 'answer'], 'Logical Reasoner', temperature=0.4, role='logical reasoner')
    heuristic_agent = LLMAgentBase(['thinking', 'answer'], 'Heuristic Solver', temperature=0.5, role='heuristic solver')

    # Get initial solutions from different agents
    numerical_thinking, numerical_answer = numerical_agent([taskInfo], numerical_instruction)
    logical_thinking, logical_answer = logical_agent([taskInfo], logical_instruction)
    heuristic_thinking, heuristic_answer = heuristic_agent([taskInfo], heuristic_instruction)

    # Collect initial answers and their thinking processes
    initial_solutions = [numerical_thinking, numerical_answer, logical_thinking, logical_answer, heuristic_thinking, heuristic_answer]

    # Initialize the Reflective Verifier agent
    reflective_verifier_agent = LLMAgentBase(['feedback', 'refined_answer'], 'Reflective Verifier Agent')

    # Perform reflective verification
    feedback, refined_answer = reflective_verifier_agent([taskInfo] + initial_solutions, reflective_verification_instruction)

    # Refine the solutions based on feedback
    refined_solutions = []
    for agent, instruction, thinking, answer in zip([numerical_agent, logical_agent, heuristic_agent], [numerical_instruction, logical_instruction, heuristic_instruction], [numerical_thinking, logical_thinking, heuristic_thinking], [numerical_answer, logical_answer, heuristic_answer]):
        refined_thinking, refined_answer = agent([taskInfo, thinking, feedback], instruction)
        refined_solutions.append(refined_thinking)
        refined_solutions.append(refined_answer)

    # Final decision-making based on all refined solutions
    final_decision_agent = LLMAgentBase(['thinking', 'answer'], 'Final Decision Agent', temperature=0.1)
    final_thinking, final_answer = final_decision_agent([taskInfo] + refined_solutions, 'Please provide the final answer based on all the reviewed and refined solutions.')
    return final_answer
\end{lstlisting}

\begin{lstlisting}[caption={Learning Agent (H6)}]
def Learning_Agent(self, taskInfo):
    # Instruction for initial reasoning
    initial_instruction = 'Please think step by step and then solve the task.'

    # Instruction for reflecting on feedback and improving the strategy
    reflect_instruction = 'Given the feedback from previous attempts, refine your strategy and solve the task again.'

    # Initialize the Learning Agent
    learning_agent = LLMAgentBase(['thinking', 'answer'], 'Learning Agent')

    # Memory to store previous attempts and feedback
    memory = []

    # Maximum number of refinement iterations
    max_iterations = 3

    for i in range(max_iterations):
        # If memory is empty, perform the initial attempt
        if not memory:
            response = learning_agent([taskInfo], initial_instruction)
        else:
            # Reflect on previous feedback and refine the strategy
            response = learning_agent([taskInfo] + memory, reflect_instruction)

        thinking, answer = response

        # Get feedback for the current attempt
        feedback_agent = LLMAgentBase(['feedback'], 'Feedback Agent')
        feedback = feedback_agent([taskInfo, thinking, answer], 'Please review the answer and provide feedback.')[0]

        # Store the current attempt and feedback in memory
        memory.extend([thinking, answer, feedback])

    # Return the final answer after all iterations
    return answer
\end{lstlisting}
  \subsection{System Prompt Format}
\label{app:system_prompt}
 
Each agent turn is conditioned on a system prompt composed of a role description and a structured output instruction.
For GSM8K, the output fields require integer answers.
The prompt is constructed as follows:
 
\vspace{4pt}
\colorbox{gray!6}{\begin{minipage}{\dimexpr\linewidth-2\fboxsep\relax}
\ttfamily\scriptsize\raggedright
\textbf{Role:} You are a \{role\}.\medskip
 
\textbf{Format instruction:}\smallskip
 
Reply EXACTLY with the following JSON format.\newline
\{"thinking": "Your thinking.", "answer": "Your answer. Return ONLY an integer. DO NOT return anything other than the integer answer."\}\newline
DO NOT MISS ANY REQUEST FIELDS and ensure that your response is a well-formed JSON object!\newline
You are to return ONLY a valid JSON object. Requirements:\newline
1) Double quotes for all keys and string values.\newline
2) No trailing commas, comments, or extra text before/after the JSON.\newline
3) Numbers must be finite (no NaN/Infinity).\newline
4) Return exactly one JSON object.\newline
Example of correct format (note that this is just an example): \{"thinking": "your thinking", "answer": 2\}
\end{minipage}}
 
\vspace{4pt}
\noindent The output schema varies by module as shown in Table~\ref{tab:modules}.
\section{Hyperparameters and Training Details}
\label{app:hyperparams}
 
All runs use the same base infrastructure built on VeRL~\citep{sheng2024hybridflow} with GRPO~\citep{shao2024deepseekmath} and turn-level ratio aggregation~\citep{wan2025rema}.
Table~\ref{tab:hyperparams_runs} details the per-run configuration for all single-controller baselines and CALM variants.
 
\begin{table}[h]
\centering
\caption{Shared hyperparameters across all GSM8K runs.}
\label{tab:hyperparams_runs}
\small
\begin{tabular}{@{}ll@{}}
\toprule
\textbf{Hyperparameter} & \textbf{Value} \\
\midrule
\multicolumn{2}{@{}l}{\textit{Model}} \\
Base model              & Llama-3.2-3B-Instruct~\citep{meta2024llama32} \\
LoRA rank / alpha       & 64 / 32 \\
LoRA target modules     & all-linear \\
\midrule
\multicolumn{2}{@{}l}{\textit{Data}} \\
Training data           & GSM8K~\citep{cobbe2021training}, 2{,}048 samples \\
Train batch size        & 32 \\
Max prompt length       & 2{,}048 tokens \\
Max response length     & 1{,}536 tokens \\
\midrule
\multicolumn{2}{@{}l}{\textit{RL objective}} \\
Algorithm               & Turn-level GRPO~\citep{shao2024deepseekmath, wan2025rema} \\
Clip mode / agg.\ mode  & turn / trajectory \\
Clip ratio $\varepsilon$    & 0.2 \\
Group size $n$          & 8 \\
KL loss coefficient     & 5e-4 \\
Entropy coefficient     & 0.01 \\
\midrule
\multicolumn{2}{@{}l}{\textit{Optimiser}} \\
Learning rate           & 5e-6 \\
LR scheduler            & constant \\
Gradient clip           & 1.0 \\
Total training steps    & 192 \\
Total epochs            & 3 \\
\bottomrule
\end{tabular}
\end{table}
 
\textsc{calm-adaptive} uses additional adaptive weighting parameters: EMA decay $\rho{=}0.9$, balancing strength $\alpha{=}1.0$, clip bounds $[w_{\min}, w_{\max}]{=}[0.1, 10.0]$, floor $\delta{=}10^{-6}$.

\section{Training Plots}
\label{app:training_plots}

\begin{figure}[!htb]
    \centering
    \includegraphics[width=\linewidth]{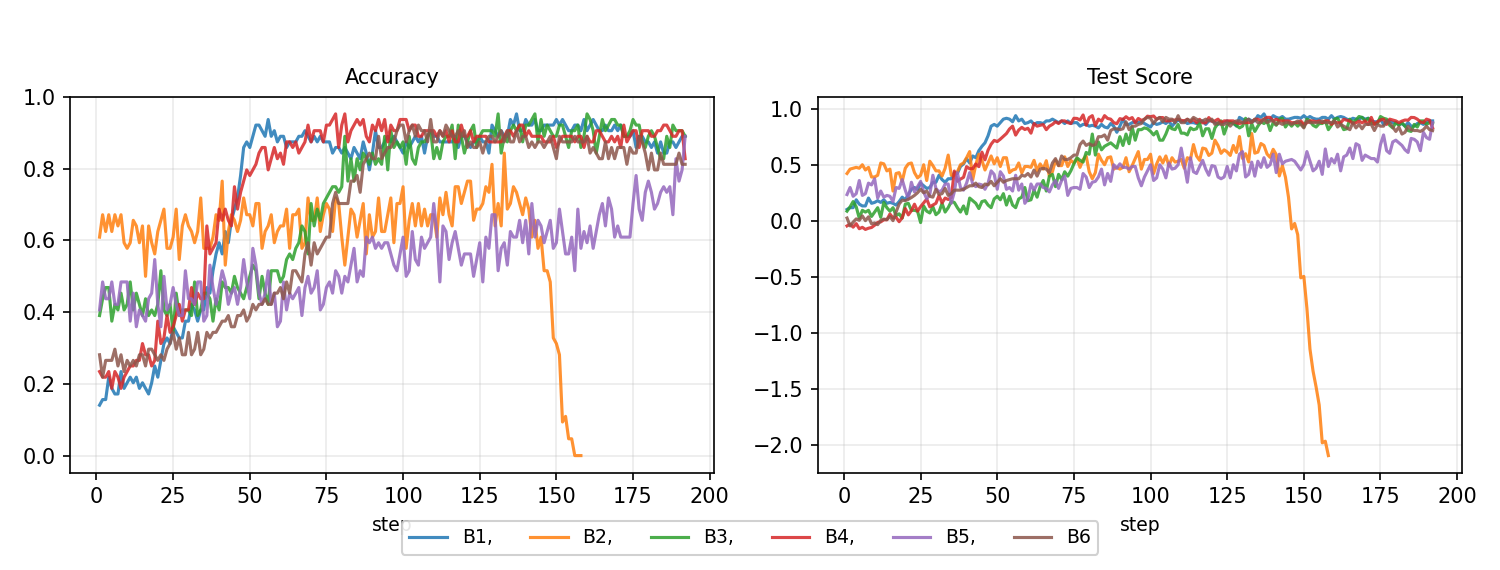}
    \caption{Training curves for different training configurations for single-controller training (B1--B6)}
    \label{fig:single_acc_plot}
\end{figure}
Figure~\ref{fig:single_acc_plot} plots the accuracy and reward score (accuracy + format reward) on a held-out validation set over training steps. We note that, under the same compute budget and hyperparameters, multi-turn controllers take longer to converge. Furthermore, Self-Refine (B2) training collapses after a few steps. For B2, we therefore evaluate the checkpoint with the best validation accuracy.

\clearpage
\section{Qualitative Examples}
\label{app:qualitative}

\subsection{Quality-Diversity: Final Decision Agent recovers from unanimous CoT failure}
\label{app:qual_debate}
 
The following trajectory ($R=1$) shows controller B5 (Quality-Diversity) on a GSM8K problem.
All three \textsc{CoT} agents misread the question as a displacement problem and apply the Pythagorean theorem; the Final Decision identifies the shared error and recovers the correct answer of 230.
 \clearpage
\begin{mdframed}[
  linecolor=black!25,
  linewidth=0.6pt,
  backgroundcolor=gray!4,
  roundcorner=4pt,
  innerleftmargin=10pt,
  innerrightmargin=10pt,
  innertopmargin=10pt,
  innerbottommargin=10pt,
  skipabove=4pt,
  skipbelow=4pt
]
 
{\small\textbf{Controller:} Quality-diversity (B5) \quad\textbf{Score:} 1.0 \quad\textbf{Turns:} 4 \quad\textbf{Ground truth:} 230}
 
\vspace{3pt}\noindent\rule{\linewidth}{0.3pt}\vspace{3pt}
 
{\small\textbf{Question}}
 
\vspace{2pt}
\colorbox{gray!8}{\begin{minipage}{\dimexpr\linewidth-2\fboxsep\relax}
\ttfamily\scriptsize
\{"question": "Two trains leave San Rafael at the same time. They begin traveling westward, both traveling for 80 miles. The next day, they travel northwards, covering 150 miles. What's the distance covered by each train in the two days?"\}
\end{minipage}}
 
\vspace{4pt}\noindent\rule{\linewidth}{0.3pt}\vspace{4pt}
 
{\small\textbf{Turn 1 --- Chain-of-Thought Agent}} \hfill {\small\textcolor{red!60!black}{answer: 174 \texttimes}}
 
\vspace{2pt}
\colorbox{gray!6}{\begin{minipage}{\dimexpr\linewidth-2\fboxsep\relax}
\ttfamily\scriptsize
\{"thinking": "Since the trains travel for 80 miles westward and then 150 miles northward, we can consider this as a right-angled triangle problem. The distance covered by each train in the two days can be calculated using the Pythagorean theorem. The total distance covered by each train is the hypotenuse of the triangle, which is the square root of (80\^{}2 + 150\^{}2). The distance covered by each train is sqrt(80\^{}2 + 150\^{}2) = sqrt(6400 + 22500) = sqrt(30400) = 174.58 miles. Since the answer should be an integer, we can round down to 174 miles for each train.", "answer": $174$\}
\end{minipage}}
 
\vspace{6pt}
 
{\small\textbf{Turn 2 --- Chain-of-Thought Agent}} \hfill {\small\textcolor{red!60!black}{answer: 260 \texttimes}}
 
\vspace{2pt}
\colorbox{gray!6}{\begin{minipage}{\dimexpr\linewidth-2\fboxsep\relax}
\ttfamily\scriptsize
\{"thinking": "Another way to solve this problem is by realizing that the total distance covered by each train forms the hypotenuse of a right-angled triangle. We can use the concept of Pythagorean triplets to simplify the calculation. The numbers 80 and 150 can be expressed as 16*5 and 25*6 respectively. Using these values, we can form a Pythagorean triplet (16*5, 25*6, 26*5) which is a multiple of the basic 3-4-5 Pythagorean triplet. The hypotenuse of this triangle is 26*5 = 130 miles. However, this is the total distance covered by each train in two days, not the individual distances covered in each direction. To find the individual distances, we can use the fact that the ratio of the sides in a Pythagorean triplet is constant. So, the ratio of the sides in our triplet is 5:6:26/5. We can use this ratio to find the individual distances covered by each train in each direction. The distance covered westward is 80 = 5*16 = 5*(5*16/26) = 5*5*16/26 = 100/13 * 5 = 500/13 miles. The distance covered northward is 150 = 6*25 = 6*(25*6/26) = 6*25*6/26 = 900/13 miles. However, these values are not integers. To get integer values, we can multiply the entire ratio by 13. The distance covered westward is 100*13/13 = 100 miles. The distance covered northward is 900/13*13 = 900 miles. But this is not the correct answer as the total distance covered in two days is 130*13 = 1690 miles, not 1690 miles = 130 miles (west) + 900 miles (north). The correct answer is 130 miles (west) + 130*6 miles (north) = 130*7 miles. So the correct answer is 130 miles (west) + 910 miles (north) = 1040 miles. But this is not the correct answer as the total distance covered in two days is 130 miles (west) + 130 miles (north) + 130 miles (east) + 130 miles (south) = 130*4 miles. So the correct answer is 130 miles (west) + 130 miles (north) = 260 miles.", "answer": $260$\}
\end{minipage}}
 
\vspace{6pt}
 
{\small\textbf{Turn 3 --- Chain-of-Thought Agent}} \hfill {\small\textcolor{red!60!black}{answer: 349 \texttimes}}
 
\vspace{2pt}
\colorbox{gray!6}{\begin{minipage}{\dimexpr\linewidth-2\fboxsep\relax}
\ttfamily\scriptsize
\detokenize{{"thinking": "We can also consider the problem as two separate right-angled triangles, one for the westward and northward distances covered on the first day, and another for the westward and northward distances covered on the second day. The first triangle has sides of 80 miles and 150 miles, and the second triangle has sides of 80 miles and 150 miles. The distance covered by each train in the two days is the sum of the distances covered in each triangle. The distance covered in the first triangle is sqrt(80^2 + 150^2) = sqrt(6400 + 22500) = sqrt(30400) = 174.58 miles. The distance covered in the second triangle is sqrt(80^2 + 150^2) = sqrt(6400 + 22500) = sqrt(30400) = 174.58 miles. The total distance covered by each train is 174.58 + 174.58 = 349.16 miles. However, we can simplify this calculation by realizing that the two triangles are congruent. Therefore, the distance covered by each train in the two days is simply 2 * 174.58 = 349.16 miles. But we need to round down to 349 miles as the answer should be an integer.", "answer": 349}}
\end{minipage}}
 
\vspace{4pt}\noindent\rule{\linewidth}{0.3pt}\vspace{4pt}
 
{\small\textbf{Turn 4 --- Final Decision Agent}} \hfill {\small\textcolor{teal!60!black}{answer: 230 \checkmark}}
 
\vspace{2pt}
\colorbox{teal!8}{\begin{minipage}{\dimexpr\linewidth-2\fboxsep\relax}
\ttfamily\scriptsize
\{"thinking": "All the solutions consider the Pythagorean theorem to calculate the total distance covered by each train. However, the correct approach is to realize that the two trains travel the same distance in each direction, so the total distance covered by each train is simply 80 miles (west) + 150 miles (north) = 230 miles. The answer should be 230 miles for each train.", "answer": 230\}
\end{minipage}}
 
\end{mdframed}
 
\vspace{4pt}
\noindent The Final Decision agent receives all three CoT outputs as context and identifies that they share the same erroneous framing,  which shows the error correcting capabilities of controller aware language models.

\end{document}